\documentclass[runningheads]{llncs}

\usepackage{eccv}

\usepackage{eccvabbrv}

\usepackage{graphicx}
\usepackage{booktabs}
\usepackage[table]{xcolor}
\usepackage{booktabs,multirow}
\usepackage[normalem]{ulem} 
\usepackage{fontawesome5}
\usepackage{marvosym}
\usepackage{amssymb}
\usepackage{xcolor}

\usepackage[accsupp]{axessibility}  

\newcommand{\cmark}{\checkmark}
\newcommand{\xmark}{$\times$}
\newcommand{\ours}{\textsc{Hi-DREAM}}

\usepackage{hyperref}

\usepackage{orcidlink}

\begin{document}

\title{Hi-DREAM: Brain-Inspired \uline{Hi}erarchical \uline{D}iffusion for fMRI-to-Image Reconstruction via \uline{R}OI \uline{E}ncoder and Visu\uline{A}l \uline{M}apping} 

\titlerunning{Hi-DREAM for fMRI-to-Image Reconstruction}

\author{Guowei Zhang\inst{1}\orcidlink{0009-0003-9309-2185} \and
Yun Zhao\inst{1,3}\orcidlink{0000-0003-2047-0029} \and
Kai Sun\inst{1}\orcidlink{0000-0002-1425-1738} \and
Moein Khajehnejad\inst{2}\orcidlink{0000-0002-2185-4596} \and
Adeel Razi\inst{2}\orcidlink{0000-0002-0779-9439} \and
Dinh Phung\inst{1}\orcidlink{0000-0002-9977-8247} \and
Levin Kuhlmann\inst{1}\orcidlink{0000-0002-5108-6348}\thanks{\Letter\ Corresponding author.}}

\authorrunning{G. Zhang et al.}

\institute{Dept of Data Science and AI, Monash University, Melbourne, Australia \\
\email{\{guowei.zhang, levin.kuhlmann\}@monash.edu} \and
School of Psychological Sciences, Monash University, Melbourne, Australia \and
Brain and Mind Centre, University of Sydney, Sydney, Australia \\
}

\maketitle

\begin{abstract}
Reconstructing natural images from fMRI requires bridging neural activity with both the structural and semantic representations used by modern generative models. Existing diffusion-based decoders often condition on a single global fMRI embedding, which limits their ability to exploit the hierarchical organization of the visual cortex and makes the contribution of different visual areas difficult to inspect. 
We propose \textbf{\ours{}}, a brain-inspired hierarchical diffusion framework that structures fMRI conditioning according to early, middle, and late visual Regions of Interest (ROI) streams. A ROI adapter converts these streams into a multi-scale cortical pyramid, and a lightweight ROI-conditioned ControlNet injects the resulting anatomy-aware priors into matched U-Net depths during denoising. 
Experiments on the Natural Scenes Dataset (NSD) show that \ours{} achieves state-of-the-art high-level semantic reconstruction while retaining strong low-level structure. Further ablation and attribution analyses show that the proposed hierarchy-aware conditioning is effective, and that different ROI streams provide
complementary, inspectable contributions to reconstruction. Our source code is available at \url{https://github.com/Zhang-gw97/Hi-Dream}.

\end{abstract}


\section{Introduction}

Reconstructing natural images from human brain activity is a central problem in visual decoding, with potential implications for understanding perception, cognition, and brain--AI alignment. Functional Magnetic Resonance Imaging (fMRI)~\cite{brown2014mri, mcrobbie2006mri} provides non-invasive access~\cite{huettel2004functional} to distributed cortical responses~\cite{ashburner2000voxel, fischl2012freesurfer}, and recent diffusion-based generative models have substantially improved the visual quality of fMRI-to-image reconstruction~\cite{takagi2023high, quan2024psychometry, lin2022mind}. However, despite this progress, a key challenge remains: most existing methods compress fMRI signals into a single global embedding before conditioning a generative model~\cite{scotti2024mindeye2, wang2024mindbridge, chen2024cinematic, chen2023seeing}. While such embeddings are effective for semantic guidance, they largely ignore the hierarchical organization of the visual cortex and make it difficult to inspect how different cortical regions contribute to different aspects of the reconstruction~\cite{nishimoto2011reconstructing, naselaris2009bayesian, shen2019deep}.

\begin{figure*}[t]
  \centering
  \includegraphics[width = \textwidth]{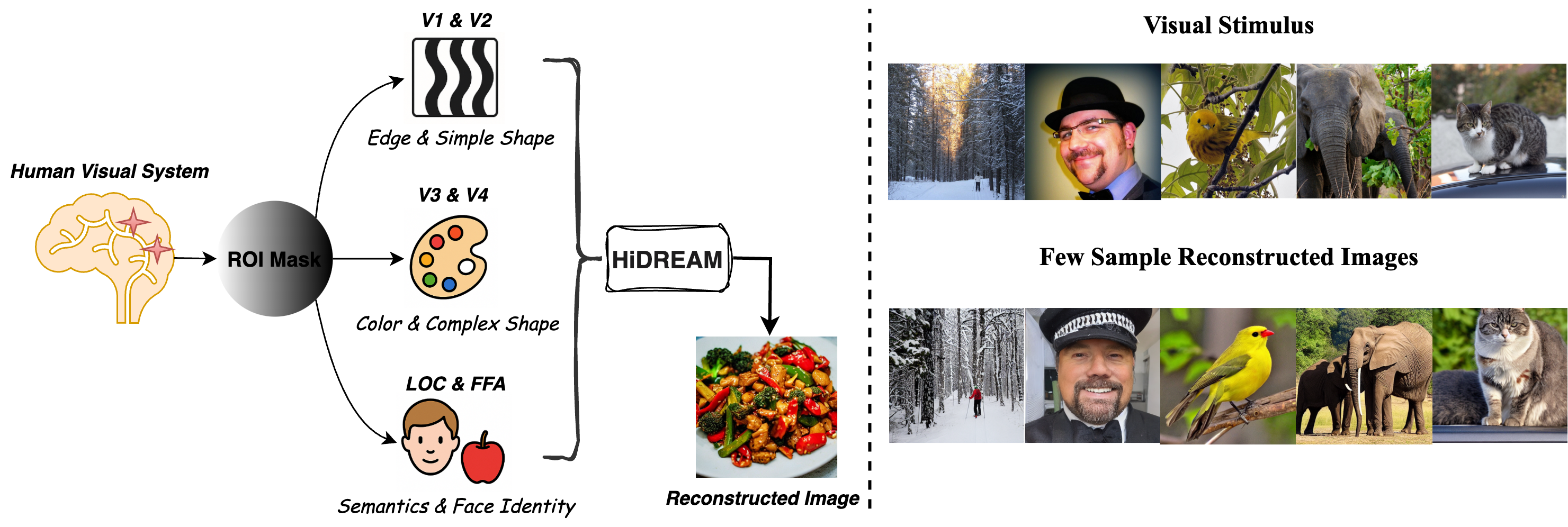}
  \caption{\textbf{Brain-inspired decoding pipeline.}
    \ours{} mirrors the visual hierarchy: fMRI signals are grouped into early/mid/late ROIs (e.g., V1/V2 for edges, V3/V4 for color and parts, LOC/FFA for semantics) and transformed by a ROI adapter into a multi-scale cortical pyramid. Right: visual stimuli and few-sample reconstructions from \ours{}, illustrating faithful structure and semantics while retaining interpretability and efficiency via compact ROI maps.}
  \label{fig:abstract}
\end{figure*}

Human visual processing is not monolithic~\cite{haxby2001distributed}. Early visual areas are closely associated with local and retinotopic structural information~\cite{joo2024brainstreams, jiang2021attention}, mid-level areas integrate contours and object parts, and late ventral regions are more related to category-level semantics (See Fig.~\ref{fig:abstract}). This hierarchy suggests that different fMRI signals may be most useful at different stages of image generation~\cite{luo2024multimodal, zhu2024adaptively}. In latent diffusion models, the U-Net backbone also contains a natural hierarchy of feature resolutions and denoising depths, where shallow layers tend to preserve spatial structure and deeper layers increasingly support semantic refinement. These observations motivate a more structured form of fMRI conditioning: instead of using one undifferentiated neural representation, the decoder should organize cortical signals into interpretable streams and inject them into matched generative depths.

We propose \textbf{\ours{}}, a brain-inspired hierarchical diffusion framework for fMRI-to-image reconstruction. \ours{} groups fMRI responses into early, middle, and late visual Regions of Interest (ROI) streams and uses a ROI adapter to convert them into a multi-scale cortical pyramid. A lightweight ROI-conditioned ControlNet~\cite{zhang2023adding}  then injects these anatomy-aware priors~\cite{mindtuner2024, throughtheir_eyes2023, neuropictor2024, qu2024taskonomy, kazmierczak2022study} into corresponding U-Net depths during denoising within a latent diffusion backbone~\cite{rombach2022high, ho2020ddpm}. This design allows low-, mid-, and high-level cortical signals to guide the generative process where they are most relevant, while avoiding the computational overhead of conditioning on full 3D fMRI volumes. The resulting framework is compact, depth-aware, and interpretable: each conditioning stream has a clear cortical correspondence, and its influence can be inspected across reconstruction stages.

We evaluate \ours{} on the Natural Scenes Dataset (NSD)~\cite{allen2022massive}, reporting results on subjects 1, 2, 5, and 7, who completed the full scanning schedule. Quantitatively, \ours{} achieves state-of-the-art (SoTA) performance on high-level semantic metrics, including Inception and CLIP, while retaining strong low-level structural fidelity. Qualitatively, the reconstructions preserve object identity, scene layout, and salient visual attributes more consistently than prior diffusion-based decoders. Beyond reconstruction performance, our ablation and attribution analyses show that the proposed hierarchy-aware conditioning is effective and interpretable: early, middle, and late ROI streams provide complementary cues; random or reversed hierarchy-depth assignments degrade performance; and the learned depth-wise contributions align with the intended cortical-to-generative hierarchy.
Our \textbf{contributions} are summarized as follows:
\begin{itemize}
    \item We propose \textbf{\ours{}}, a brain-inspired hierarchical diffusion framework that structures fMRI conditioning according to early, middle, and late visual ROI streams.
    \item We design a \textbf{ROI adapter} and \textbf{depth-matched ROI-ControlNet} that convert cortical responses into compact multi-scale priors and inject them into corresponding U-Net depths.
    \item We demonstrate on NSD that \ours{} achieves \textbf{SoTA} semantic reconstruction while retaining strong low-level structure, with ablation showing that the ROI streams contribute complementary and inspectable information.
\end{itemize}


\section{Related Work}

\subsection{fMRI-based image reconstruction.} 

A growing line of work decodes natural images from fMRI by coupling neural representations with powerful image generators~\cite{takagi2023high, aggarwal2024ensemble, li2024enhancing}. Early diffusion-era methods such as Mind-Vis~\cite{chen2023seeing} map fMRI features into latent diffusion models, while MindEye2~\cite{scotti2024mindeye2}, Psychometry~\cite{quan2024psychometry}, DREAM~\cite{xia2024dream}, and MindBridge~\cite{wang2024mindbridge} improve semantic alignment, perceptual quality, or cross-subject transfer through stronger encoders and generator-side conditioning. Recent diffusion-based decoders further explore structured guidance: MindDiffuser~\cite{lu2023minddiffuser} uses decoded semantic and structural features, Animate Your Thoughts~\cite{lu2025animate} extends reconstruction to video by disentangling semantic, structural, and motion factors, and NeuralDiffuser~\cite{li2025neuraldiffuser} introduces primary-visual gradient guidance for fine-detail fidelity. Most of these methods condition the generator using a compact global fMRI embedding or decoded visual features.

\subsection{Conditional diffusion and control.}
Modern diffusion models~\cite{rombach2022high, vaswani2017attention, li2021align} admit a spectrum of conditioning mechanisms, including cross-attention to token-like sequences, feature injection at selected layers, and auxiliary networks that modulate intermediate activations~\cite{ho2020ddpm,rombach2022high, ho2022classifier}. ControlNet~\cite{zhang2023adding} exemplifies the latter, adding a lightweight branch trained to translate structured conditions (edges, depth, poses) into spatial guidance for the U-Net with minimal changes to the sampler. For fMRI decoding, prior works mainly condition on global or low-resolution neural embeddings that act analogously to text tokens~\cite{takagi2023high, scotti2024mindeye2, wang2024mindbridge, quan2024psychometry}, which offers semantic steering but provides limited spatial specificity. Using structured spatial conditions as guidance remains comparatively under-explored in this domain.

\subsection{Neuroanatomical priors and ROI modeling.}
Neuroimaging and systems neuroscience establish a hierarchical organization of human visual cortex~\cite{bandettini2009neuroimaging, baillet2017meg, cohen2017eeg}: early areas (V1/V2/V3) encode local, retinotopic structure, mid-level regions integrate contours/parts, and late ventral stream areas represent categorical semantics~\cite{GrillSpectorMalach2004,DiCarloZoccolanRust2012,Kravitz2013VentralStream,WandellWinawer2015Retinotopy}. Computational decoders that explicitly respect this hierarchy are rare; most pipelines flatten the neural signal into a single embedding~\cite{takagi2023high,scotti2024mindeye2}, obscuring how different regions contribute. A few works~\cite{kay2013compressive, finn2015functional, huth2016natural} inject ROI information as features~\cite{grill2014functional, benson2018human} or masks~\cite{wandell2007visual}, but rarely model cooperative interactions between ROI groups across network depths. In contrast, we partition ROIs into early/mid/late groups with group-specific transformations and couple them to the generator via a depth-matched, ControlNet-based injection~\cite{zhang2023adding}, supplying spatial priors that localize structure. Together these choices yield a decoder that is not only competitive in accuracy but also more neuro-aligned than purely data-driven alternatives.


\section{Method}
\label{sec:method}

\subsection{Motivation and Overview}
\label{sec:overview}

\textbf{Motivation.}
Human visual cortex is hierarchically organized: early areas encode local and retinotopic structure, mid-level areas integrate contours and object parts, and late ventral regions represent category-level semantics. We use this organization as an inductive bias for fMRI-to-image reconstruction. Instead of mapping all fMRI activity into a single global embedding, we structure the conditioning signal into early, middle, and late ROI streams and inject them into matched depths of a diffusion backbone.

\begin{figure*}[t]
  \centering
  \includegraphics[width = \textwidth]{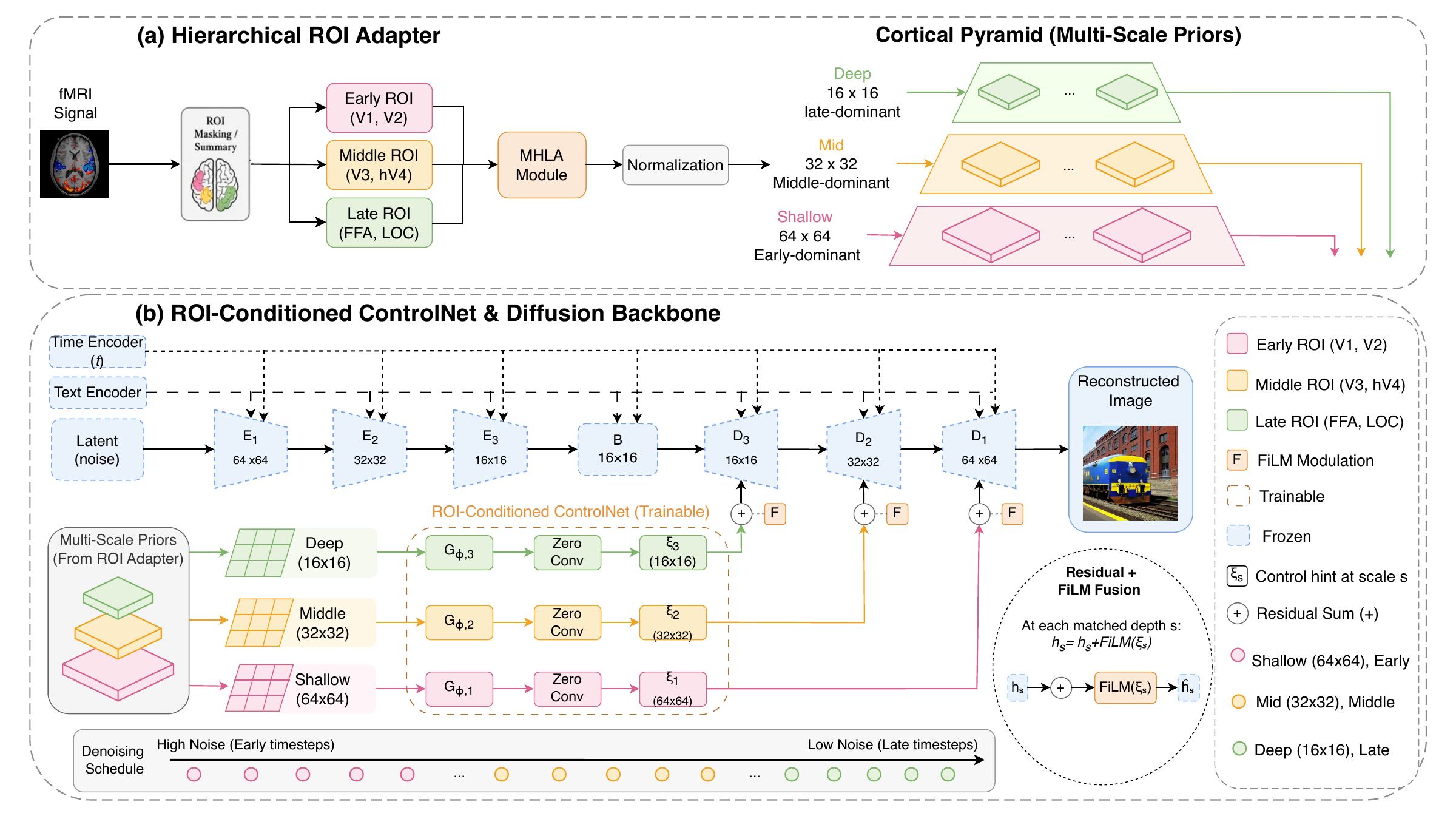}
    \caption{\textbf{Overview of \ours.}
    (a) The ROI adapter constructs early, middle, and late ROI streams from fMRI responses and converts them into a multi-scale cortical pyramid, with MHLA coordinating cross-stream interactions.
    (b) The ROI-conditioned ControlNet injects the resulting anatomy-aware priors into matched U-Net depths through residual--FiLM modulation, guiding diffusion reconstruction without full-volume fMRI conditioning and enabling model-level attribution.}
  \label{fig:model}
\end{figure*}

\textbf{Overview.}
As shown in Fig.~\ref{fig:model}, \ours{} contains two lightweight components. First, a ROI adapter (Sec.~\ref{sec:adapter}) converts trial-wise fMRI responses and subject-specific ROI masks into a multi-scale cortical pyramid. Second, a ROI-conditioned ControlNet (Sec.~\ref{sec:controlnet}) translates this pyramid into depth-matched hints and injects them into the corresponding U-Net blocks during denoising. Given ROI responses and masks, the adapter produces scale-wise priors $\{\Xi_s\}_{s\in\mathcal{S}}$, each $\Xi_s$ is converted into a hint $\xi_s$ and injected through residual $+$ FiLM modulation. The latent diffusion backbone then denoises under these depth-specific hints to generate the reconstructed image. During inference (Sec.~\ref{sec:train}), we record group weights and injected responses for model-level attribution.

\subsection{ROI Stream Construction}
\label{sec:roi_stream}

Let the fMRI input be $\mathbf{X}\in\mathbb{R}^{T\times H\times W\times D}$, and let $\{\mathcal{R}^e,\mathcal{R}^m,\mathcal{R}^l\}$ denote the early, middle, and late ROI groups. For each ROI $r$, we compute a regional response $a_r\in[0,1]$ from the normalized trial-wise beta amplitude and use a subject-specific projected ROI mask $m_r\in\{0,1\}^{H\times W}$ aligned to the image plane. These quantities provide the activity strength and anatomical support for constructing ROI condition maps.

\textbf{Scale-space construction.}
We build a spatial pyramid at resolutions $\mathcal{S}=\{64,32,16\}$ to match the shallow, middle, and deep U-Net blocks. At scale $s$, we define a soft ROI mask: 
\begin{equation}
\label{equ:roi_adapter}
    \tilde{m}_{r,s}=\mathrm{Down}_{s}(\mathcal{G}_{\sigma_s} * m_r)\in[0,1]^{h_s\times w_s},
\end{equation}
where $\mathcal{G}_{\sigma_s}$ is a Gaussian kernel and $\mathrm{Down}_{s}$ resizes the mask to the corresponding U-Net resolution. The blurred mask itself does not encode stimulus-specific edges; instead, it provides anatomy-constrained spatial scaffolding. Layout and detail preservation arise from the combination of trial-wise ROI activation, shallow depth-matched guidance, and the learned denoising prior.

\textbf{Per-scale weighting and grouping.}
Each ROI contributes a trial-specific condition map
\begin{equation}
    c_{r,s}=w_{r,s}\,a_r\,\tilde{m}_{r,s},
\end{equation}
where $w_{r,s}\in[0,1]$ is a learnable depth-dependent gate. Stacking maps within a ROI group $g\in\{e,m,l\}$ yields a group tensor $\mathbf{C}_{g,s}\in\mathbb{R}^{R_g\times h_s\times w_s}$.

\subsection{Hierarchical ROI Adapter}
\label{sec:adapter}

The ROI adapter converts the group tensors $\{\mathbf{C}_{g,s}\}$ into compact, scale-wise cortical priors for the diffusion model.

\textbf{Lightweight channelization.}
A $1\!\times\!1$ Conv/MLP mixer $M_{g,s}$ compresses each group tensor $\mathbf{C}_{g,s}$ into guidance channels $\Xi_{g,s}\in\mathbb{R}^{d_s\times h_s\times w_s}$, where $d_s$ is small in practice, e.g., $d_s\in\{8,16,32\}$. The group-specific guidance maps are then combined with learnable nonnegative mixing weights $\alpha_{g,s}$ to form the scale-wise cortical pyramid:
\begin{equation}
\label{equ:lightweight}
    \Xi_{s}\;=\;\sum_{g\in\{e,m,l\}}\alpha_{g,s}\,\Xi_{g,s},\qquad s\in\mathcal{S}.
\end{equation}
Equation~\ref{equ:lightweight} allows each U-Net scale to receive a different mixture of early, middle, and late ROI evidence, supporting depth-aware conditioning rather than using a single fixed fMRI representation across all layers.

\textbf{Multi-Head Latent Attention.}
The ROI streams are not treated as independent modules. After group-specific features are obtained, we use a Multi-Head Latent Attention (MHLA)~\cite{vaswani2017attention} module to model cooperative interactions among early, middle, and late streams. U-Net features act as queries, while ROI-derived latents provide keys and values. The resulting gated attention features modulate the depth-specific ControlNet hints, allowing information from one ROI group to influence how other groups are used at a given denoising depth. Thus, the ROI-ControlNet provides spatial, depth-matched guidance, while MHLA coordinates cross-ROI interactions.

\textbf{Stability and regularization.}
To stabilize the learned mixtures, we constrain the ROI gates $w_{r,s}$ and group weights $\alpha_{g,s}$ to be nonnegative and normalize them across groups at each scale. We further apply mild regularization to avoid degenerate single-stream solutions and use a weak anatomical prior that encourages shallow scales to rely more on early ROIs and deeper scales to rely more on late ROIs, while still allowing the final mixing weights to be learned from data.

\subsection{ROI-Conditioned ControlNet: Selective Depth Injection}
\label{sec:controlnet}

The ROI-conditioned ControlNet translates the adapter pyramid $\{\Xi_s\}$ into resolution-aligned hints that guide image generation. Rather than using ROI features as generic conditioning, we inject each hint into the U-Net depth where the corresponding scale is most informative.

\textbf{Resolution-aligned injection.}
For each scale $s$ matched to a U-Net block with activation $h_s$, a shallow stack $G_{\phi,s}$ of $3{\times}3$ convolutions produces a hint $\xi_s=G_{\phi,s}(\Xi_s)$. We inject it using residual-FiLM fusion:
\begin{equation}
\label{equ:injection}
    \hat h_s \;=\; h_s \;+\; \lambda_s\Big(A_s(\xi_s)\;+\;\gamma_s(\xi_s)\odot h_s\;+\;\beta_s(\xi_s)\Big),
\end{equation}
where $A_s$ is a $1{\times}1$ convolution, $(\gamma_s,\beta_s)$ are per-channel affine parameters, and $\lambda_s\in[0,1]$ controls the guidance strength at depth $s$.

\textbf{Hint normalization and calibration.}
We normalize each hint to zero mean and unit variance, and bound the total injected energy by a soft budget:
\begin{equation}
    \sum_s \lambda_s \cdot \mathbb{E}\!\left[\|A_s(\xi_s)\|_1\right]\le \eta,
\end{equation}
implemented by on-the-fly rescaling of $\{\lambda_s\}$. This prevents any single scale from overwhelming the denoising backbone.

\textbf{Depth and time scheduling.}
We encourage early, middle, and late ROI evidence to contribute more strongly at shallow, intermediate, and deep U-Net blocks, respectively. Guidance is also scheduled over diffusion time:
\begin{equation}
    \lambda_s(t)=\lambda_s^{\max}\cdot \tfrac{1}{2}\big(1+\cos(\pi\cdot\mathrm{clip}(t/T,0,1))\big)^{\rho_s},
\end{equation}
so structural hints can guide early denoising while deeper semantic hints refine the reconstruction later.

\subsection{Training and Inference}
\label{sec:train}

\textbf{Objective.}
We optimize a diffusion objective augmented with lightweight structural regularization:
\begin{equation}
\label{equ:loss}
    \mathcal{L}_{\text{total}} = \mathcal{L}_{\text{diff}} + \lambda_1\mathcal{L}_1 + \lambda_2\mathcal{L}_{\text{grad}} + \lambda_3(1-\mathrm{SSIM}),
\end{equation}
where $\mathcal{L}_1$, $\mathcal{L}_{\text{grad}}$, and $1-\mathrm{SSIM}$ encourage pixel consistency, edge/gradient preservation, and structural similarity, respectively. We additionally retain regularizers on $\alpha$ and $w$ to stabilize ROI mixing and prevent single-stream dominance.

\textbf{Two-stage schedule.}
In Stage~A, we freeze the diffusion backbone and warm up the ROI adapter to produce stable cortical pyramids. The mixing weights $\alpha_{g,s}$ are initialized conservatively, and $\lambda_s$ is ramped up from small values to avoid unstable early conditioning. In Stage~B, we enable the ROI-conditioned ControlNet and selectively unfreeze only the U-Net blocks aligned with each scale. The ControlNet learns to translate $\{\Xi_s\}$ into effective guidance signals, while the partially unfrozen backbone adapts locally and preserves the pretrained generative prior. All experiments use the training objective in Eq.~\ref{equ:loss}, together with depth-wise guidance $(\lambda_{\text{shallow}},\lambda_{\text{mid}},\lambda_{\text{deep}})=(1.2,0.6,0.3)$.

\textbf{Inference and complexity.}
At test time, $\{\Xi_s\}$ is computed in one pass and DDIM/PLMS sampling proceeds with fixed $(\alpha_{g,s},\lambda_s)$. The adapter and ControlNet operate on compact 2D pyramids, adding negligible overhead relative to the U-Net and avoiding the I/O cost of repeatedly conditioning on full 3D fMRI volumes.


\section{Experiment}
\label{sec:exp}

\subsection{Dataset and Metrics}
\label{sec:dataset}

\textbf{Dataset.}
We evaluate our method on the Natural Scenes Dataset (NSD)~\cite{allen2022massive}, a large-scale 7T fMRI dataset with natural-image stimuli. NSD provides trial-wise beta responses and subject-specific anatomical/ROI information, including retinotopic visual areas and higher-level category-selective regions. Following common practice in fMRI-to-image reconstruction, we report results on subjects 1, 2, 5, and 7, who completed the full scanning schedule. We group the visual ROIs into early areas (V1/V2), middle areas (V3/hV4), and late ventral areas (LOC/FFA), which serve as the cortical streams used by \ours{}.

\textbf{Evaluation Metrics.}
We follow the standard evaluation protocol used in prior fMRI reconstruction work~\cite{scotti2024mindeye2, wang2024mindbridge} and report metrics covering low-level, perceptual, and semantic fidelity. 
Low-level fidelity is evaluated using PixCorr~\cite{ren2023reconstructing}, SSIM~\cite{wang2004image} and AlexNet(2/5)~\cite{krizhevsky2012imagenet} ($\uparrow$), which measure pixel-, structure-, and early perceptual-level similarity. High-level semantic alignment is evaluated using Inception~\cite{szegedy2016rethinking} and CLIP~\cite{radford2021learning} ($\uparrow$), together with EfficientNet-B distance~\cite{tan2019efficientnet} and SwAV distance~\cite{caron2020unsupervised} ($\downarrow$), to assess category-level consistency and feature-level discrepancy with the target stimulus.
For fair comparison with previous methods, all reconstructions are evaluated at $256{\times}256$ resolution using the same preprocessing and metric definitions across baselines.

\subsection{Qualitative \& Quantitative Results}
\label{sec:comparison}

\textbf{Main results.}

Table~\ref{tab:recon-results} compares \ours{} with previous state-of-the-art (SoTA) fMRI-to-image reconstruction methods, including Takagi~\etal{}~\cite{takagi2023high}, MindBridge~\cite{wang2024mindbridge}, DREAM~\cite{xia2024dream}, Psychometry~\cite{quan2024psychometry}, and MindEye2~\cite{scotti2024mindeye2}.

\begin{table}[htbp]
  \centering
  \caption{Quantitative comparison between \ours{} and previous SoTA fMRI-to-image reconstruction methods on low-level and high-level metrics. All results are averaged across subjects 1, 2, 5, and 7 from the Natural Scenes Dataset. Best results are shown in \textbf{bold}, and second-best results are \underline{underlined}.}
  
  \label{tab:recon-results}
  \begin{tabular*}{\textwidth}{@{\extracolsep{\fill}} l c c c c  c c c c}
    \toprule
    
    \multirow{2}{*}{\textbf{Methods}}
      & \multicolumn{4}{c}{\textbf{Low-Level}}
      & \multicolumn{4}{c}{\textbf{High-Level}} \\
    
    \cmidrule(lr){2-5} \cmidrule(lr){6-9}
      & PixCorr$\uparrow$ & SSIM$\uparrow$ 
      & Alex(2)$\uparrow$ & Alex(5)$\uparrow$ 
      & Incep$\uparrow$   & CLIP$\uparrow$  
      & Eff$\downarrow$   & SwAV$\downarrow$  \\
    
    \midrule
    Takagi et al.~\cite{takagi2023high}
      & .246     & \underline{.410}     & 78.9\% & 85.6\%
      & 83.8\% & 82.1\% & .811      & .504  \\ 
    
    MindBridge~\cite{wang2024mindbridge}
      & .148  & .259  & 86.9\% & 95.3\%
      & 92.2\% & 94.3\% & .713   & .413  \\
    
    Psychometry~\cite{quan2024psychometry}
      & .297  & .340  & \underline{96.4}\% & \textbf{98.6\%}
      & \underline{95.8}\% & \underline{96.8}\% & .628   & .345  \\

    DREAM~\cite{xia2024dream}
      & .274  & .328  & 93.9\% & 96.7\%
      & 93.4\% & 94.1\% & .645   & .418  \\
    
    MindEye2~\cite{scotti2024mindeye2}
      & \textbf{.322}  & \textbf{.431}  & 96.1\% & \textbf{98.6\%}
      & 95.4\% & 94.5\% & \underline{.619}   & \underline{.344} \\
    
    \midrule
    
    \cellcolor{blue!10}\textbf{\ours{}} &
    \cellcolor{blue!10}\underline{.316} & 
    \cellcolor{blue!10}.407 & 
    \cellcolor{blue!10}\textbf{96.8\%} & 
    \cellcolor{blue!10}\underline{97.4}\% & 
    \cellcolor{blue!10}\textbf{97.2\%} & 
    \cellcolor{blue!10}\textbf{98.3\%} & 
    \cellcolor{blue!10}\textbf{.541} & 
    \cellcolor{blue!10}\textbf{.342} \\
    
    \bottomrule
  \end{tabular*}
\end{table}

\ours{} achieves the strongest high-level semantic alignment, obtaining the best results on Incep, CLIP, Eff, and SwAV. Qualitatively, this means our reconstructions capture object identity and scene semantics more reliably, while avoiding the mode-mismatch sometimes observed when conditioning directly on full fMRI volumes. 
It also performs strongly on low-level and perceptual metrics, achieving the best AlexNet(2) score and second-best PixCorr and AlexNet(5) scores. 
These results suggest that hierarchy-aware ROI conditioning improves semantic reconstruction while preserving competitive structural and perceptual fidelity.

\begin{figure}[ht]
  \centering
  \includegraphics[width = \textwidth]{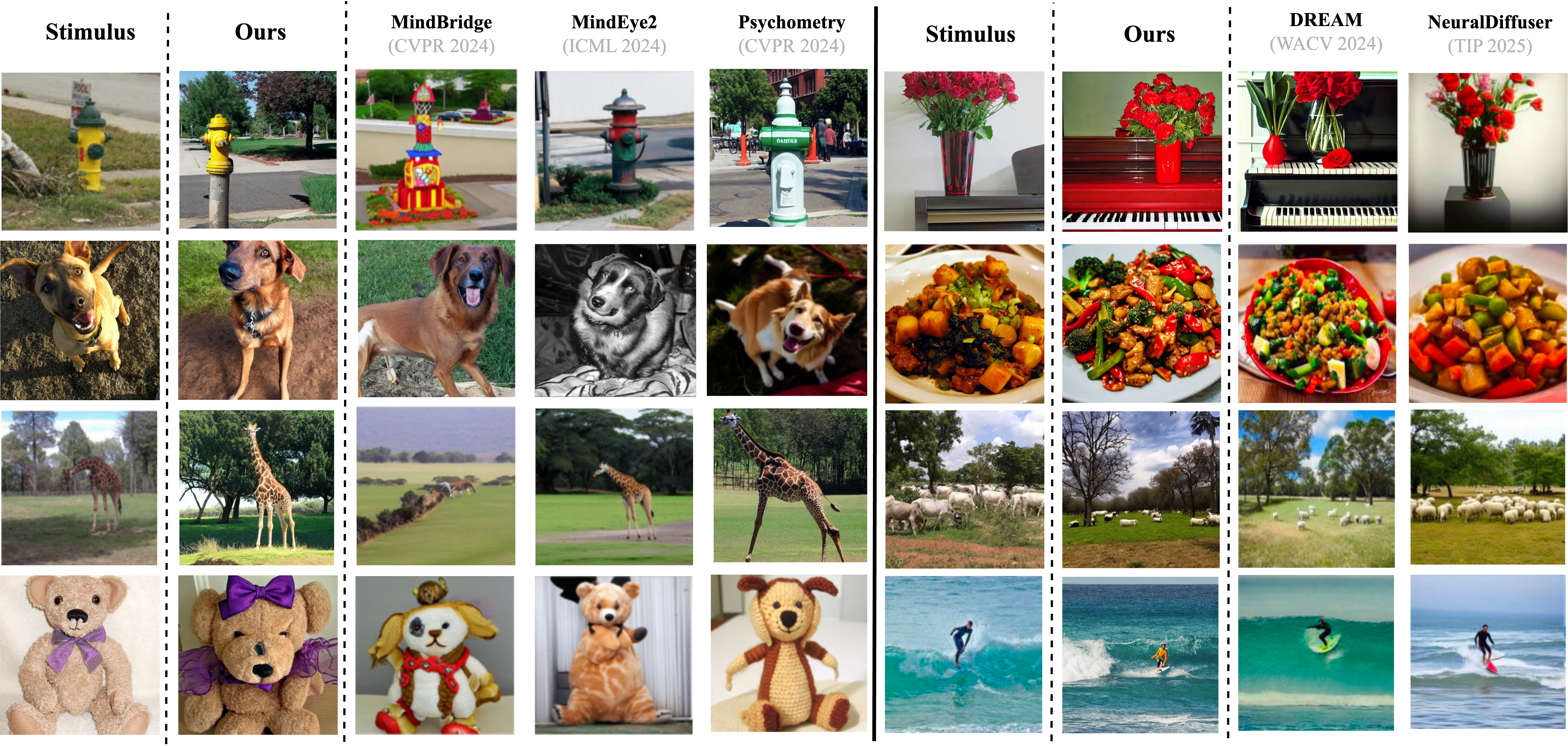}
  \caption{Qualitative comparison on the NSD test set (NSD Shared 1000). Each group shows the stimulus (left) and the corresponding reconstructions by \ours{} and prior methods.}
  \label{fig:image_result}
\end{figure}

Fig.~\ref{fig:image_result} further shows that our reconstructions achieve consistently higher semantic correctness than prior methods. Across diverse stimuli, our results preserve the correct category and salient attributes, with notably better color fidelity (e.g., the red flower bouquet and piano scene, the characteristic color palettes in food images) and more accurate shape/geometry (object silhouettes, part proportions, and pose). We also retain more coherent scene layout and context, reducing common failure modes in baselines such as color drift, category confusion, or structural collapse. Overall, our reconstructions are closer to the intended content of the stimulus, indicating stronger high-level semantic alignment.

\textbf{Why \ours{} helps.}
Compared with prior methods, two design choices appear decisive: 
(i) a brain-inspired ROI adapter that separates early/mid/late groups and builds a multi-scale cortical pyramid. This aligns the information each group provides with the depth of the U-Net (layout-sensitive structure at shallow layers, parts/semantics at deeper layers), reducing representational frictions that arise when a single embedding is forced to serve all depths. 
(ii) a selective, depth-matched ControlNet that injects ROI evidence where it is most useful, with per-scale strength control. This avoids over-conditioning and prevents shallow detail from drowning out deeper semantic cues. 
Together, these choices yield stronger semantic metrics without substantially increasing computation: the adapter and ControlNet operate on compact subject-specific maps rather than full 3D fMRI volumes, improving data throughput and simplifying optimization.

\textbf{Relation to specific baselines.}
MindBridge~\cite{wang2024mindbridge} focuses on cross-subject knowledge transfer; Psychometry~\cite{quan2024psychometry} optimizes perceptual and semantic objectives with powerful priors; MindEye2~\cite{scotti2024mindeye2} integrates fMRI encoders with diffusion via cross-attention. \ours{} is orthogonal and complementary: instead of seeking a single best fMRI embedding, we structure the conditioning by cortical hierarchy and scale, which preserves interpretability (each signal maps to an interpretable ROI stream) while driving gains on semantically grounded metrics. The consistent advantage on Inception/CLIP and feature distances, averaged over the four full-coverage NSD subjects, supports this interpretation.

\subsection{Ablation Study}
\label{sec:ablation}

\textbf{(A) Composition of ROI streams.}

We first examine how each cortical stream contributes to reconstruction. The ROI adapter separates fMRI signals into Early, Middle, and Late streams, corresponding to retinotopic structure, intermediate shape/part information, and category-selective semantics. Rather than treating these streams as isolated predictors, \ours{} combines them through a shared diffusion decoder and MHLA, allowing their contributions to interact across denoising depths. Table~\ref{tab:abl_a1} reports the reconstruction quality when using Early only, Early+Middle, and all three streams.

\begin{table*}[ht]
    \centering
    \caption{ROI-stream ablation showing complementary contributions of Early, Middle, and Late streams.}
    \label{tab:abl_a1}
    \begin{tabular*}{\textwidth}{@{\extracolsep{\fill}} l c c c c c c c c}
        \toprule
        ROI Streams & PixCorr$\uparrow$ & SSIM$\uparrow$ & Alex(2)$\uparrow$ & Alex(5)$\uparrow$ & Incep$\uparrow$ & CLIP$\uparrow$ & Eff$\downarrow$ & SwAV$\downarrow$ \\
        \midrule
        E & .267 & \textbf{.421} & 92.1\% & 94.0\% & 92.4\% & 92.1\% & .628 & .431 \\
        E+M & .294 & .402 & 95.1\% & 96.5\% & 95.5\% & 96.6\% & .572 & .363 \\
        E+M+L & \textbf{.316} & .407 & \textbf{96.8\%} & \textbf{97.4\%} & \textbf{97.2\%} & \textbf{98.3\%} & \textbf{.541} & \textbf{.342} \\
        \bottomrule
    \end{tabular*}
\end{table*}

Early-only conditioning achieves strong structural fidelity, especially on SSIM, suggesting that early retinotopic ROIs provide useful spatial scaffolding. Adding Middle improves both low-level and semantic metrics, indicating that mid-level ROIs bridge local structure and part-level information. 

The full Early + Middle + Late model obtains the best overall balance, with the highest PixCorr, AlexNet, Inception, and CLIP scores and the lowest feature distances. These results suggest that the streams contribute complementary interacting biases rather than independent linear effects.

To test whether the proposed hierarchy is an arbitrary grouping, we compare against two controls with the same model capacity in Table~\ref{tab:abl_a2}. Random assigns ROIs randomly to the three streams, while Reversed injects early visual ROIs into deeper layers and late ROIs into shallower layers. Both controls degrade performance, especially Reversed on high-level metrics, indicating that the depth–ROI correspondence is important for effective conditioning.

\begin{table*}[ht]
    \centering
    \caption{Hierarchy-control experiments with the same model configuration. Random and reversed ROI-depth assignments degrade performance, indicating the importance of matching cortical hierarchy to U-Net depth.}
    \label{tab:abl_a2}
    \begin{tabular*}{\textwidth}{@{\extracolsep{\fill}} l c c c c c c c c}
        \toprule
        Setting & PixCorr$\uparrow$ & SSIM$\uparrow$ & Alex(2)$\uparrow$ & Alex(5)$\uparrow$ & Incep$\uparrow$ & CLIP$\uparrow$ & Eff$\downarrow$ & SwAV$\downarrow$ \\
        \midrule
        Random & .281 & .356 & 92.7\% & 95.1\% & 94.3\% & 95.6\% & .589 & .381 \\
        Reversed & .264 & .331 & 90.8\% & 93.6\% & 91.7\% & 93.2\% & .628 & .414 \\
        \textbf{Ours} & \textbf{.316} & \textbf{.407} & \textbf{96.8\%} & \textbf{97.4\%} & \textbf{97.2\%} & \textbf{98.3\%} & \textbf{.541} & \textbf{.342} \\
        \bottomrule
    \end{tabular*}
\end{table*}

\textbf{(B) Role of the ROI Adapter and MHLA.}

We next isolate the contribution of the hierarchical adapter and the MHLA module. The ROI adapter converts subject-specific cortical activity into multi-scale anatomy-aware priors, while MHLA coordinates interactions among Early, Middle, and Late streams before depth-specific injection. This separation allows us to test whether performance gains come from spatial ROI conditioning alone, or from explicitly modeling cross-ROI cooperation.

\begin{table}[ht]
    \centering
    \small
    \caption{Ablation of hierarchy modules. The ROI Adapter provides anatomy-aware spatial priors, while MHLA further improves cross-stream coordination and semantic alignment.}
    \resizebox{\linewidth}{!}{
        \begin{tabular}{cc|cccccccc}
        \toprule
        
        Adapter & MHLA & PixCorr$\uparrow$ & SSIM$\uparrow$ & Alex(2)$\uparrow$ & Alex(5)$\uparrow$ & Incep$\uparrow$ & CLIP$\uparrow$ & Eff$\downarrow$ & SwAV$\downarrow$ \\
        
        \midrule
        
        \xmark & \xmark & .096 & .191 & 76.8\% & 79.4\% & 84.0\% & 83.5\% & .718 & .543 \\
        \cmark & \xmark & .205 & .302 & 84.6\% & 88.8\% & 92.2\% & 92.1\% & .626 & .487 \\
        \checkmark & \checkmark & \textbf{.316} & \textbf{.407} & \textbf{96.8\%} & \textbf{97.4\%} & \textbf{97.2\%} & \textbf{98.3\%} & \textbf{.541} & \textbf{.342} \\
        
        \bottomrule
        \end{tabular}
    }
    \label{tab:abl_b}
\end{table}

As shown in Table~\ref{tab:abl_b}, without the ROI adapter, the ControlNet branch lacks structured cortical priors and performs poorly across both low- and high-level metrics. Adding the ROI adapter improves reconstruction by supplying depth-aligned spatial guidance. 

Adding MHLA further improves both low-level fidelity and high-level semantic alignment, while also reducing feature distances, suggesting that cross-stream interaction benefits structural and semantic reconstruction simultaneously. In short, the adapter provides anatomical priors, and MHLA coordinates how these priors cooperate across U-Net depths.

\textbf{(C) Qualitative ROI-stream attribution.}

\begin{figure*}[ht]
  \centering
  \includegraphics[width=\columnwidth]{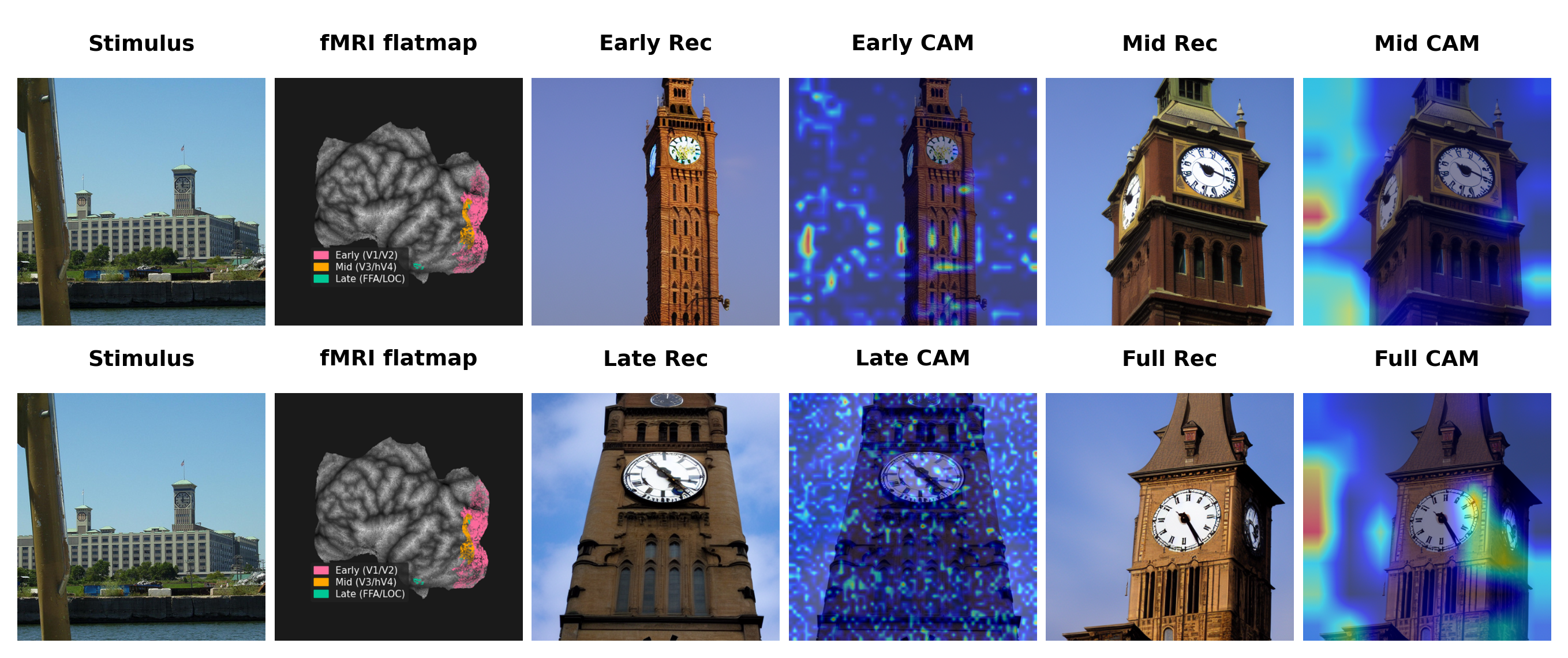}
  \caption{Qualitative ROI-stream ablation and Grad-CAM visualization. For the same stimulus and projected fMRI flatmap, we compare reconstructions generated from Early-only, Middle-only, Late-only, and Full ROI conditioning. Grad-CAM overlays show how different ROI streams induce distinct spatial attention patterns during reconstruction.}
  \label{fig:abl_c}
\end{figure*}

To complement the quantitative ablations, we visualize how different ROI streams affect the reconstruction process using Grad-CAM overlays. Fig.~\ref{fig:abl_c} shows reconstructions from the same stimulus when conditioning on Early-only, Middle-only, Late-only, and all ROI streams. 

The ROI-specific reconstructions reveal distinct visual biases. With Early-only conditioning, the model roughly preserves the global scene layout, including the tall tower-like structure and the surrounding building silhouette, but the object identity and fine architectural details remain unstable. With Middle-only conditioning, the reconstruction better captures part-level structure, such as the clock face, roof shape, and local building components, although the overall geometry is still distorted. With Late-only conditioning, the model emphasizes category-level cues of a clock tower, but the reconstruction becomes less spatially grounded and loses much of the original scene context.

The full model combines these complementary signals and produces a more balanced reconstruction, preserving both the clock-tower identity and its spatial arrangement. Its Grad-CAM overlay is also more spatially organized, with strong responses around visually salient regions such as the foreground railing/occlusion and the clock face. 
These regions are also naturally informative for human recognition of the scene, suggesting that the full model attends to semantically meaningful and structurally important parts of the image. This qualitative analysis is consistent with the quantitative ablations: individual ROI streams induce different reconstruction biases, whereas their combination provides a more coherent structural-semantic reconstruction.

\textbf{(D) Depth-wise ROI contribution.}

We further examine whether Hi-DREAM uses each ROI stream at the intended U-Net depth. For ROI group $g\in\{\text{Early},\text{Middle},\text{Late}\}$ and depth $s$, we measure the normalized injected energy
\begin{equation}
    \mathcal{I}_{g,s}
    =
    \mathbb{E}_{x,t}
    \left[
    \left\|
    \alpha_{g,s}\,\Xi_{g,s}(x,t)
    \right\|_1
    \right],
\end{equation}
where $\Xi_{g,s}$ is the group-specific map before aggregation and $\alpha_{g,s}$ is the learned group weight. We row-normalize $\mathcal{I}_{g,s}$ across ROI groups at each depth and average over the test set, so larger values indicate stronger reliance on that ROI stream.

\begin{figure}[ht]
  \centering
  \includegraphics[width=0.9\columnwidth]{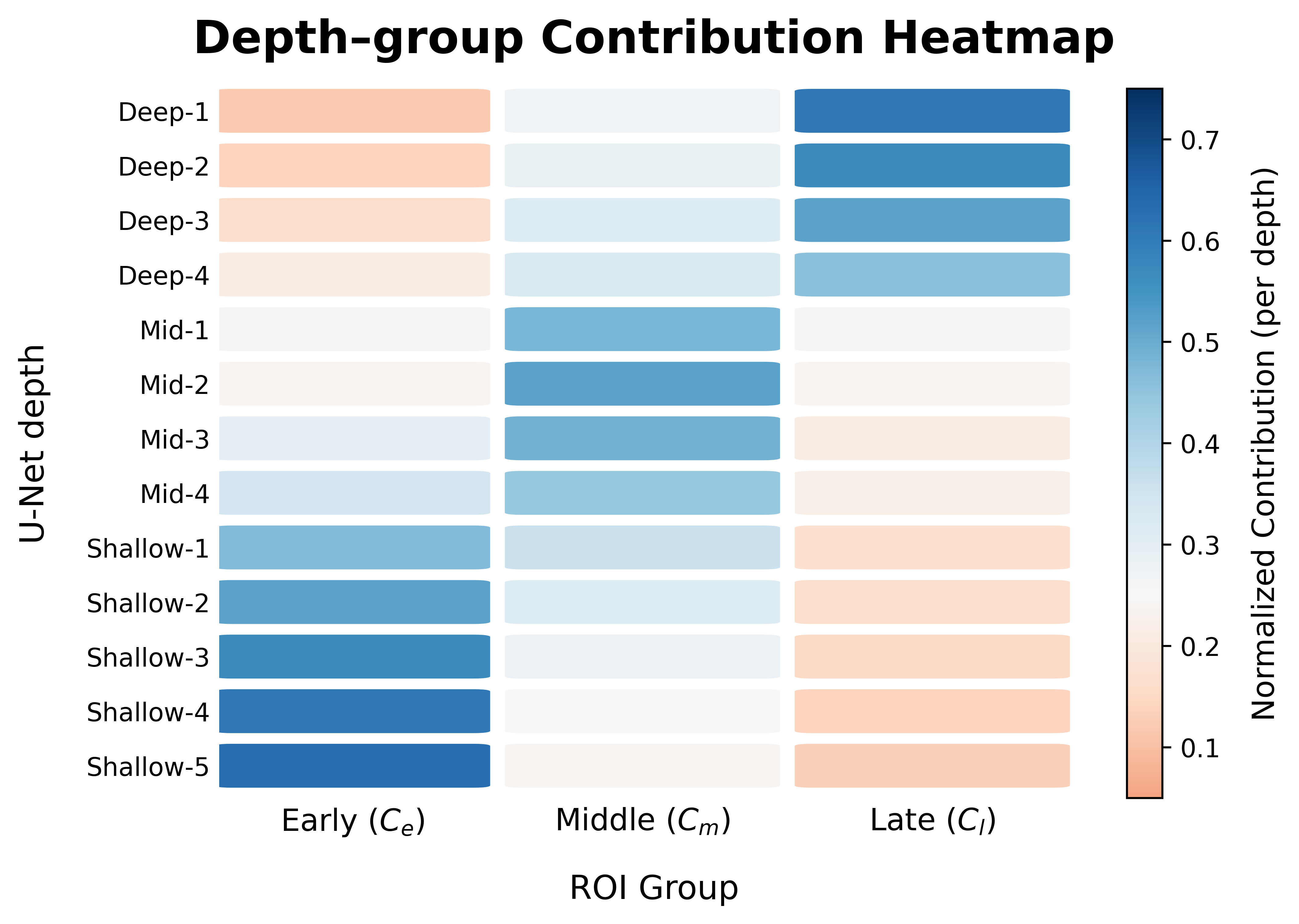}
  \caption{Depth--ROI alignment measured by normalized injected energy. Rows are ordered from deep to shallow U-Net blocks; darker blue indicates higher relative contribution.}
  \label{fig:abl_d}
\end{figure}

As shown in Fig.~\ref{fig:abl_d}, the learned contribution pattern follows the intended hierarchy: 
Late, Middle, and Early streams show stronger relative contributions at deep, intermediate, and shallow blocks, respectively. Since the injection sites are fixed but the group weights and gates are learned from data, this pattern indicates that the model does not simply use ROI maps as generic conditions. Instead, Hi-DREAM learns a depth-consistent use of cortical streams, supporting the design that early ROIs provide spatial scaffolding, middle ROIs bridge part-level structure, and late ROIs refine semantic content.

\subsection{Discussion}
\label{sec:discussion}

\textbf{Complementary roles of ROI streams.}
The ablation results show that the benefit of \ours{} does not come simply from adding more conditioning parameters. Early-only conditioning provides strong structural fidelity, adding middle ROIs improves part-level and semantic consistency, and the full early-middle-late configuration achieves the best overall balance. The random and reversed hierarchy controls further show that the correspondence between ROI group and U-Net depth is important for effective reconstruction. These findings support the view that different visual ROI streams provide complementary cues that should be integrated in a depth-aware manner.

\textbf{Interpretability.}
Beyond performance, \ours{} provides an inspectable conditioning pathway. Since each stream corresponds to a defined cortical ROI group and each condition is injected at a known U-Net depth, we can analyze how different cortical priors influence the reconstruction process. The Grad-CAM visualizations and depth-wise contribution heatmap show that ROI streams induce different spatial attention patterns and that their learned contributions follow the intended early/middle/late hierarchy. These analyses should be interpreted as model-level evidence of how the decoder uses ROI information, rather than direct evidence of biological mechanisms.

\textbf{Efficiency and practicality.}
The proposed design is also computationally practical. By converting fMRI responses into compact 2D cortical pyramids, \ours{} avoids the cost of repeatedly conditioning on full 3D fMRI volumes while preserving subject-specific anatomical information. This makes the framework suitable for iterative experimentation and leaves room for future extensions, such as incorporating broader visual ROIs, learning cross-subject canonical spaces, or adapting the hierarchy-aware conditioning strategy to other brain decoding tasks.

\subsection{Limitations}

Although \ours{} achieves strong semantic reconstruction and improved structural fidelity, fine-grained visual details remain challenging. In particular, small textures, object boundaries, and facial configurations can still be imperfect when the fMRI signal is weak or when semantic guidance dominates local structure. Improving such details while preserving the current semantic alignment is an important direction for future work.

Another limitation is anatomical coverage. \ours{} focuses on a compact set of representative early, middle, and late visual ROIs, which provides a clean hierarchical design but does not capture the full richness of visual cortex. Additional ventral, dorsal, motion-sensitive, and body-selective regions may provide complementary cues for object layout, motion, pose, and category-specific reconstruction.

Finally, the current framework is trained in a subject-specific manner and uses subject-specific ROI masks. This design allows precise anatomical conditioning, but it also limits direct transfer across individuals. Future work will explore shared cortical parcellations, canonical ROI spaces, and few-shot adaptation to reduce subject dependence and move toward more generalizable fMRI-to-image decoders.


\section{Conclusion}
In this paper, we introduced \ours{}, a brain-inspired diffusion framework that conditions on fMRI via a cortical hierarchy. By converting early/mid/late ROIs into a multi-scale cortical pyramid and injecting it with a depth-matched ROI-ControlNet, \ours{} provides efficient, anatomy-aware guidance and interpretable control over where structure and semantics enter the generator. On the NSD test set, \ours{} reaches SoTA performance on high-level semantic metrics while retaining strong low-level structure. Looking ahead, we aim to strengthen low-level detail without sacrificing semantics and to broaden anatomical coverage and improve cross-subject generalization, moving toward a decoder that more closely mirrors human visual processing.


\section*{Acknowledgements}
This work was supported in part by a Monash University scholarship. The authors acknowledge the computational resources provided by the Monash M3 high-performance computing platform and the National Computational Infrastructure (NCI), Australia. These resources supported the large-scale model training, inference, and experimental evaluation conducted in this study.

%
%

\bibliographystyle{splncs04}
\bibliography{main}

\end{document}